\pdfoutput=1

\documentclass[11pt]{article}

\usepackage{emnlp2021}

\usepackage{times}
\usepackage{latexsym}
\usepackage{graphicx}
\usepackage{amsfonts}
\usepackage{amsmath}
\usepackage{multirow}

\usepackage[T1]{fontenc}

\usepackage[utf8]{inputenc}

\usepackage{microtype}

\title{Integrating Deep Event-Level and Script-Level Information for \\ Script Event Prediction}

\author{Long Bai, Saiping Guan, Jiafeng Guo, Zixuan Li, Xiaolong Jin, Xueqi Cheng \\
	CAS Key Lab of Network Data Science and Technology, Institute of \\Computing Technology, Chinese Academy of Sciences (CAS); \\
	School of Computer Science and Technology, University of Chinese Academy of Sciences \\
	\texttt{\{bailong18b,guansaiping,guojiafeng,lizixuan,jinxiaolong,cxq\}@ict.ac.cn} \\
}

\begin{document}
	\maketitle
	\begin{abstract}
		Scripts are structured sequences of events together with the participants,
		which are extracted from the texts.
		Script event prediction aims to predict the subsequent event given the historical events in the script.
		Two kinds of information facilitate this task,
		namely, the event-level information and the script-level information.
		At the event level, existing studies view an event as a verb with its participants,
		while neglecting other useful properties, such as the state of the participants.
		At the script level, most existing studies only consider a single event sequence corresponding to one common protagonist.
		In this paper, we propose a Transformer-based model, called MCPredictor,
		which integrates deep event-level and script-level information for script event prediction.
		At the event level, MCPredictor utilizes the rich information in the text to obtain more comprehensive event semantic representations.
		At the script-level,
		it considers multiple event sequences corresponding to different participants of the subsequent event.
		The experimental results on the widely-used New York Times corpus demonstrate the effectiveness and superiority of the proposed model.
		
	\end{abstract}
	
	\section{Introduction}
	\label{sec:introduction}
	
	Scripts, consisting of structured sequences of events, are a kind of knowledge that describes everyday scenarios~\cite{abelson1977scripts}.
	A typical script is the \textit{restaurant script}, which describes the scenario of a person going to a restaurant.
	In this script, ``customer enter restaurant'', ``customer order food'', ``customer eat food'' and ``customer pay bill'' happen successively.
	This structured knowledge is helpful to downstream natural language processing tasks, such as anaphora resolution~\cite{bean-riloff-2004-unsupervised} and story generation~\cite{chaturvedi-etal-2017-story}.
	
	Script event prediction aims to predict the subsequent event based on the historical events in the script.
	What is vital to this task is to understand the historical events comprehensively. 
	Therefore, two categories of information are essential,
	namely, the event-level information and the script-level information.
	The event-level information contains necessary elements to describe the events, such as the verbs and their participants,
	while the script-level information describes how the events are connected and structured, 
	such as via the temporal order or a common participant.
	
	\begin{figure}
		\centering
		\includegraphics[scale=0.33]{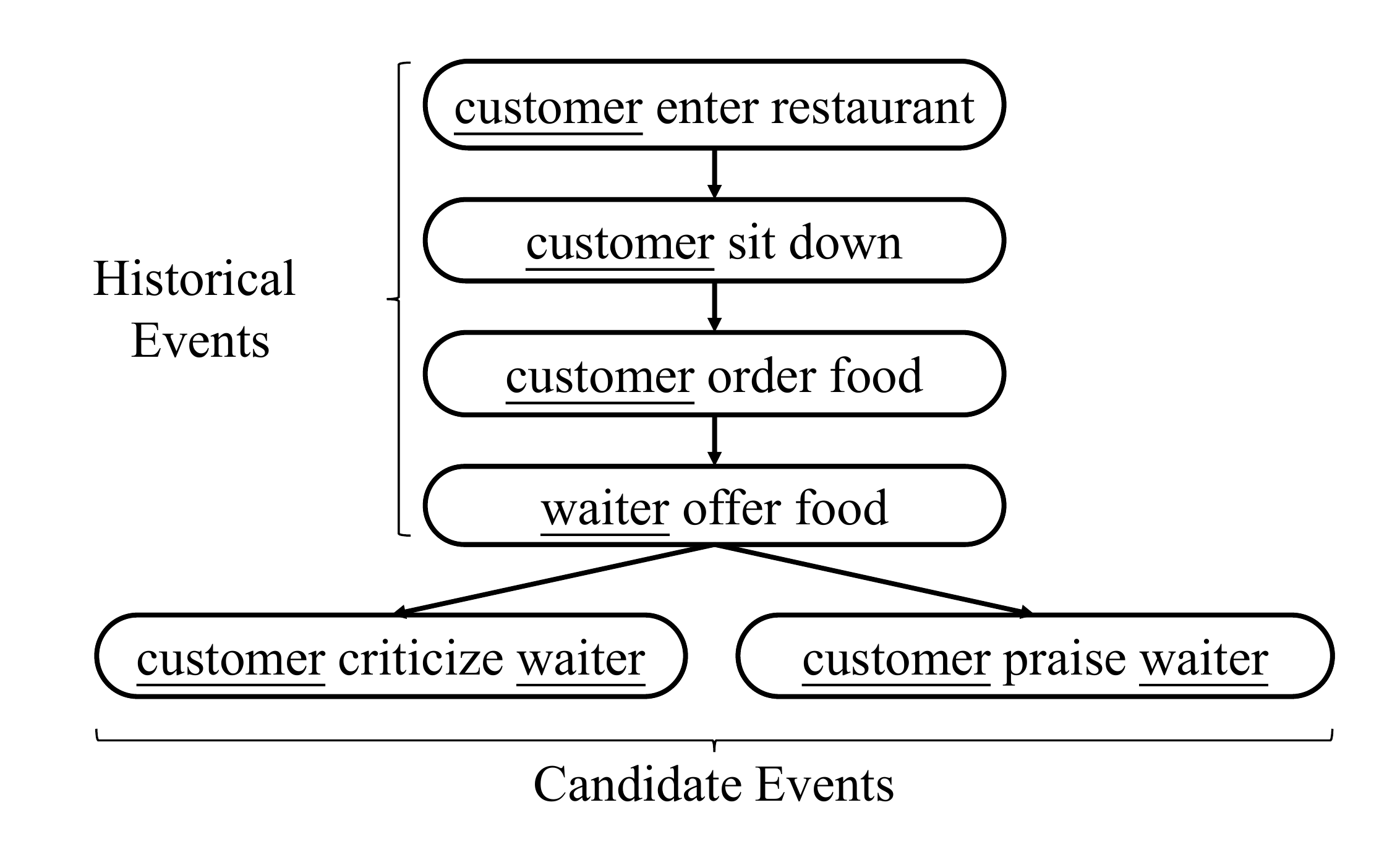}
		\caption{
			An example of the script event prediction task,
			where two essential participants are underlined.
		}
		\label{fig:introduction}
	\end{figure}
	
	The existing studies only consider the verb with its participants (usually the headwords) at the event level.
	However, this event representation method suffers from a lack of necessary information to derive a more accurate prediction.
	There exists other important properties of the events in the original texts, such as the intention and the state of the participants.
	For instance, as shown in Figure~\ref{fig:introduction},
	if the waiter's service is friendly,
	the customer will be more likely to praise the waiter.
	If the customer is irritable,
	he/she will be more likely to criticize the waiter.
	Unfortunately, the current formulation does not consider these features.
	From the aspect of the script-level information,
	existing studies only model the events that share a specific protagonist.
	These events are organized into a sequence by temporal order, which is thus called the narrative event chain.
	However, an event may contain multiple participants,
	each of which has its own influence on the occurrence of the event.
	As shown in Figure~\ref{fig:introduction},
	the two participants, i.e., the customer and the waiter,
	take their own actions and jointly determine whether the customer will criticize or praise the waiter.
	
	Motivated by these observations,
	at the event level, we trace the events back to their original texts and consider all the constituents in the texts that describe the events.
	Then, we utilize the informative constituents to obtain more comprehensive event semantic representations.
	With respect to the script level,
	we view the script as multiple narrative event chains derived from the participants of the subsequent event.
	Via modeling the narrative event chains of multiple participants,
	we are able to capture their behavior trends
	and predict the occurrence of the subsequent event.
	Thus, by integrating the above deep event-level and script-level information,
	we propose MCPredictor, a Transformer-based script event prediction model.
	At the event level, MCPredictor contains an event encoding component and a text encoding component.
	Via aggregating the output of the two components,
	it obtains more comprehensive event semantic representations.
	At the script level, MCPredictor contains a chain modeling component and a scoring component.
	The former integrates the temporal order information into event representations.
	The latter aggregates the influence of multiple narrative event chains through an attention mechanism to predict the subsequent event.
	In general, this paper has the following main contributions:
	\begin{itemize}
		\item We emphasize the importance of both the event-level information and the script-level information for script event prediction,
		and go deep into it;
		
		\item We propose the MCPredictor model to integrate both kinds of information for script event prediction. It introduces rich information from the original texts to enhance the event-level information and learns the script-level information by aggregating the influence of multiple narrative event chains on the subsequent event;
		
		\item The proposed model achieves state-of-the-art performance on the widely-used benchmark dataset. The event-level and the script-level information introduced by us are also testified to be effective.
	\end{itemize}
	
	\section{Related Work}
	
	Recent studies on script event prediction start from \cite{chambers-jurafsky-2008-unsupervised},
	which proposes the basic structure of the narrative event chain with a specific participant (called the protagonist).
	Each event is represented as a tuple of the verb and the dependency relation between the verb and the protagonist, 
	i.e., $\langle verb, dependency \rangle$.
	Then, it uses Pointwise Mutual Information (PMI) to measure the score of two events to be in the same narrative event chain.
	Finally, it aggregates the pairwise scores to infer a candidate's probability of being the subsequent event of the narrative event chain.
	
	\citet{balasubramanian-etal-2013-generating} pointed out that the event representation method mentioned above may lose the co-occurrence information between a subject and its object.
	Therefore, they represented an event as a $\langle subject, verb, object \rangle$ triple.
	\citet{pichotta-mooney-2014-statistical,AAAI1611995} further extended the event representation method by taking the indirect object into consideration.
	Following studies on script event prediction are mainly based on this event representation method.
	The above symbolic representation methods may cause the sparsity problem.
	Therefore, distributed event representation method is applied in more recent studies~\cite{DBLP:journals/corr/ModiT13,AAAI1611995}.
	In addition, early studies aggregate the scores of each event in the narrative event chain and the candidate event,
	ignoring the temporal order of the events.
	Therefore, \citet{AAAI1612157,wang-etal-2017-integrating,DBLP:conf/aaai/LvQHHH19} introduced LSTM to integrate temporal order information
	\footnote{Following \cite{wang-etal-2017-integrating}, 
		we use the term ``temporal order''. 
		Precisely, it is the ``narrative order''.}.
	\citet{ijcai2018-584} further converted the narrative event chain to a narrative event evolutionary graph and used graph neural networks to model it.
	
	Conventionally, an event contains a verb and several participants,
	where the headwords are used to represent the participants.
	This event representation method suffers from a lack of information since only a few words are considered.
	Therefore, \citet{ding-etal-2019-event-representation} used if-then commonsense knowledge to enrich the event representation.
	\citet{lee-goldwasser-2019-multi,zheng2020incorporating,lee-etal-2020-weakly} labeled extra discourse relations between the events according to the conjunctions between them.
	\citet{zheng2020incorporating,lv-etal-2020-integrating,lee-etal-2020-weakly} introduced pre-trained language models, such as BERT~\cite{devlin-etal-2019-bert}, to script event prediction and achieved excellent performance.
	Still, these studies lose some of the informative constituents in the texts that directly describe the events.
	
	The above models all focus on a single narrative event chain.
	The only study that considers the multiple chains is~\cite{chambers-jurafsky-2009-unsupervised}.
	However, our study is different from it in the following aspects: 
	(1) when representing an event, it only considers the verb and the dependency relation between the verb and the protagonist, 
	while we keep all the participants in an event;
	(2) it adopts the symbolic event representation method and a pair-based model (PMI),
	while we use distributed event representation method and consider the temporal order information.
	
	\section{Problem Statement}
	\label{sec:problem_statement}
	
	In this paper, an event $e = \langle v, a_0, a_1, a_2, t \rangle$ consists of a verb $v$, three participants
	\footnote{
		Usually, an event only contains one or two participants,
		However, to uniformly characterize all possible conditions,
		we pad the missing participants with null.
	}
	(i.e., subject $a_0$, object $a_1$, and indirect object $a_2$), and its original sentence $t$ where the event appears.
	Here, $t=\{w_0, w_1, ...\}$ consists of a sequence of words.
	
	The script event prediction task aims to predict the most probable subsequent event $e^*$ given the candidate event set $\mathcal{S}$ and the historical events $\mathcal{H}$.
	Here, $\mathcal{S}=\{e_{c_0},e_{c_1}, ..., e_{c_{m-1}}\}$ consists of the $m$ candidate events and $\mathcal{H}=\{e_0, e_1, ...\}$ consists of the happened events.
	Note that, since the candidate events have not happened yet,
	there are no corresponding sentences for them.

	\section{The MCPredictor Model}
	
	\begin{figure*}
		\centering
		\includegraphics[scale=0.45]{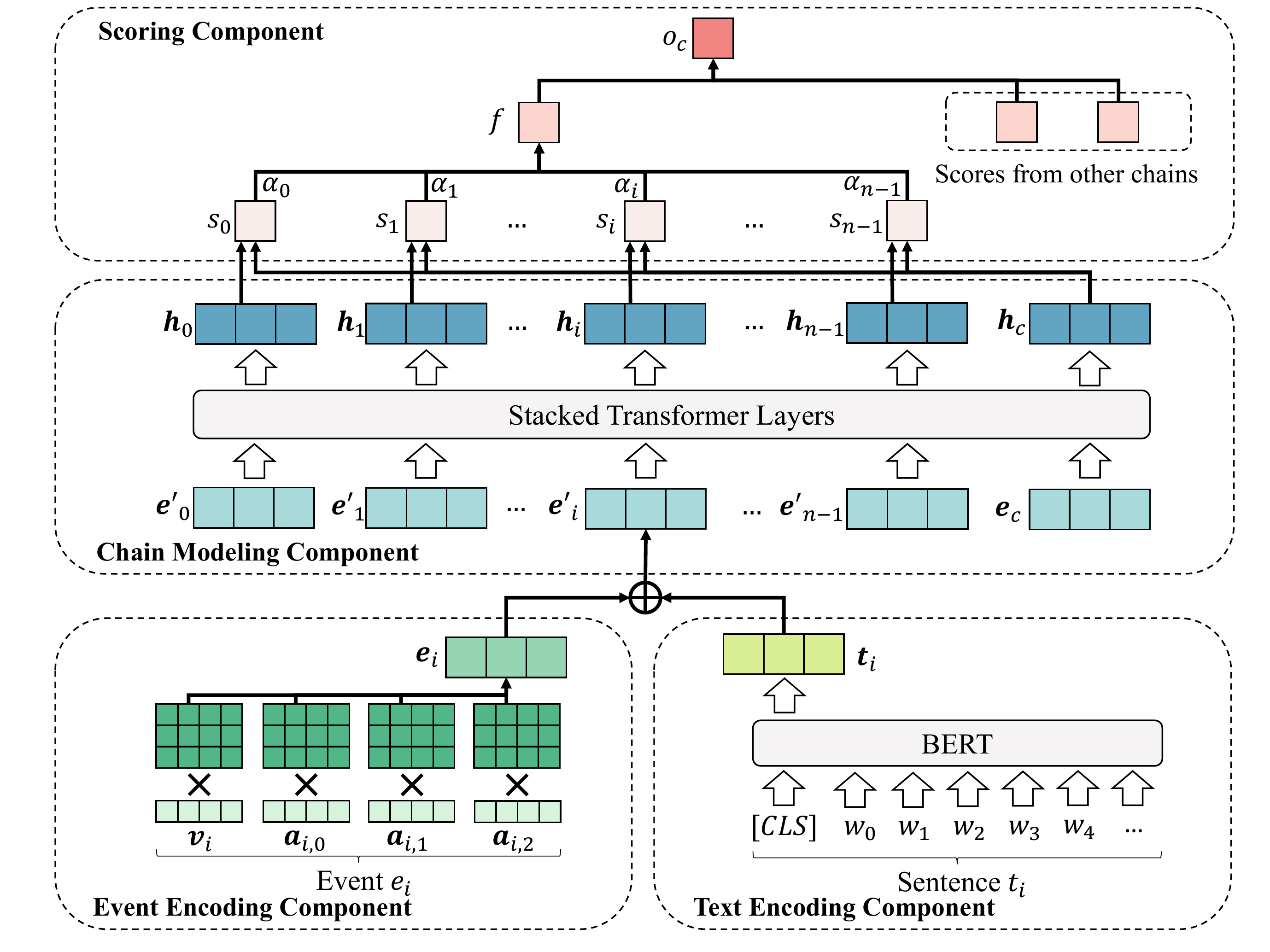}
		\caption{An illustrative diagram of the proposed MCPredictor model.}
		\label{fig:model}
	\end{figure*}
	
	In this section, we will describe our MCPredictor model.
	As show in Figure~\ref{fig:model},
	it consists of four main components:
	(1) event encoding,
	(2) text encoding,
	(3) chain modeling,
	and (4) scoring.
	We will describe the model in single-chain mode
	and then introduce how the results from the multiple chains (derived from different participants of the candidate event) are aggregated.
	Finally, we will list the variants of the scoring component.
	
	\subsection{Event Encoding Component}
	
	The event encoding component aims to map the events into low-dimensional vectors.
	The embeddings $\mathbf{e}_{i}$\footnote{
		In what follows, following the convention, the embeddings are denoted with the same letters but in boldface.
	}
	for events $e_{i}$ ($i\in\{0, ..., n-1\}$, $n$ is the number of historical events) is calculated via mapping the verbs $v_{i}$ and their participants $a_{i,0}, a_{i,1}, a_{i,2}$ into the same vector space $\mathbb{R}^{d_w}$:
	\begin{equation}
		\begin{split}
			\mathbf{e}_{i} =& 
			\tanh (W_v^T \mathbf{v}_{i} + W_{a0}^T \mathbf{a}_{i,0} + \\
			& W_{a1}^T \mathbf{a}_{i,1} + W_{a2}^T \mathbf{a}_{i,2} + \mathbf{b}_e),
		\end{split}
		\label{eq:event_embedding}
	\end{equation}
	where $W_v, W_{a0}, W_{a1}, W_{a2}\in \mathbb{R}^{d_w\times d_e}$ are mapping matrices;
	$\mathbf{b}_e \in \mathbb{R}^{d_e}$ is the bias vector;
	$d_w$ is the word embedding size and $d_e$ is the event embedding size.
	The candidate event $e_c$ is encoded similarly.
	
	\subsection{Text Encoding Component}
	
	The text encoding component is to encode the sentence where each event appears into a vector.
	Since pre-trained language models, such as BERT~\cite{devlin-etal-2019-bert},
	have shown great power in encoding natural language,
	we apply BERT-tiny in this component 
	and use the output embedding of the ``[CLS]'' tag as the sentence embedding.
	
	Specifically, to avoid information leakage,
	we replace the other events in the same narrative event chain with the ``[UNK]'' tag when encoding the sentence of the focused event in this component.
	In addition, we add role tags before and after the verb to specify dependency relation (``[subj]'', ``[obj]'' or ``[iobj]'') between the verb and the protagonist.
	For instance, the sentence \textit{``He entered the restaurant and asked the waiter for the menu''}
	will be converted into \textit{``He [subj] entered [subj] the restaurant and [UNK] ''}, if we focus on the event corresponding to the verb ``entered''.
	
	Then, the embeddings of events $e_i$ that integrate sentence information are:
	\begin{equation}
		\mathbf{e}'_{i} = \mathbf{e}_{i} + \text{BERT}(t_{i}),
		\label{eq:text_encoder}
	\end{equation}
	where BERT is the BERT-tiny network and $t_{i}$ are the converted sentence corresponding to events $e_{i}$.
	
	\subsection{Chain Modeling Component}
	\label{sec:chain_modeling}
	In the above components, 
	the events are separately encoded without temporal information.
	Therefore, we use stacked Transformer~\cite{NIPS2017_7181} layers to deeply integrate the temporal order information into the event representations.
	Following the convention~\cite{wang-etal-2017-integrating,ijcai2018-584},
	to model the interactions between the historical events (i.e., the historical narrative event chain) and the candidate event,
	we append the candidate event to the end of the historical narrative event chain.
	As mentioned in Section~\ref{sec:problem_statement},
	there is no corresponding sentence for the candidate event,
	thus, we only use its event embedding calculated by Equation~\ref{eq:event_embedding}.
	Additionally, the positional embeddings~\cite{NIPS2017_7181} are added to the event embeddings to specify their positions in the chain.
	Finally, the last Transformer layer outputs the hidden vectors $\mathbf{h}_{i}$ and $\mathbf{h}_c$ for historical events $e_{i}$ and the candidate event $e_c$, respectively.
	
	\subsection{Scoring Component}
	
	The scoring component aims to calculate the score of each candidate event.
	This component contains two steps:
	event-level scoring and script-level aggregation.
	The former calculates the similarity score between each historical event and the candidate event.
	The latter aggregates these scores to derive the similarity score between the script and the candidate event.
	Besides, we will describe the variants of this component.
	
	\subsubsection{Event-Level Scoring}
	
	Aggregating the event-level similarity scores after modeling the temporal order is better than only considering the event-pair similarity scores or only using the last hidden vector of the chain for prediction,
	as verified by \cite{wang-etal-2017-integrating}.
	Therefore, before evaluating the script-level similarity scores,
	we calculate the pairwise similarity scores $s_i$ between the candidate event $e_c$ and historical events $e_{i}$ using the hidden vectors from the chain modeling component in Section~\ref{sec:chain_modeling}.
	Here, we use the negative Euclidean distance (denoted as E-Score) as the similarity scores:
	\begin{equation}
		s_{i} = - || \mathbf{h}_{i} - \mathbf{h}_c ||_2.
	\end{equation}
	
	\subsubsection{Script-Level Aggregation}
	This step is to aggregate the event-level scores from multiple narrative event chains.
	Let us begin with a single narrative event chain.
	
	We use an attention mechanism to derive the similarity score $f$ between the historical narrative event chain and the candidate event $e_c$:
	\begin{equation}
		f = \sum_{i=0}^{n-1}\alpha_{i}s_{i},
		\label{result:singlechain}
	\end{equation}
	where the attention weight $\alpha_{i}$ of each event $e_{i}$ is calculated by:
	\begin{equation}
		\alpha_{i} = \frac{\exp(u_{i})}{\sum_{k=0}^{n-1}\exp(u_{k})},
	\end{equation}
	where we use the scaled-dot product attention~\cite{NIPS2017_7181}:
	\begin{equation}
		u_{i} = \frac{1}{\sqrt{d_e}}\mathbf{h}_{i}^T \mathbf{h}_c .
		\label{att:scaled-dot}
	\end{equation}
	
	The similarity score is calculated by aggregating the similarity scores from multiple chains:
	\begin{equation}
		o_{c_i}=\sum_{j=0}^{2}f_{i,j},
	\end{equation}
	where $f_{i,j}$ is the similarity score between the candidate event $e_{c_i}$ and the historical narrative event chain corresponding to its $j$-th participant.
	
	Then, the probability of each candidate event $e_{c_i}\in \mathcal{S}$ to be the correct subsequent event is calculated by:
	\begin{equation}
		\Pr(e_{c_i}|\mathcal{H}) = \frac{\exp(o_{c_i})}{\sum_{k=0}^{m-1}\exp(o_{c_k})},
	\end{equation}
	where $o_{c_i}$ is the score of each candidate event.
	
	Note that, in previous studies, only a single narrative event chain of the given protagonist is considered in the historical events $\mathcal{H}$.
	Therefore, the score of each candidate event $o_{c_i}$ is equal to $f_{i}$, where $f_{i}$ represents the similarity score of the historical narrative event chain and the candidate event $e_{c_i}$ calculated via Equation~\ref{result:singlechain}.
	
	Finally, we select the most possible candidate event as the subsequent event $e^*$ as follows:
	\begin{equation}
		e^* = \arg\max_{e_{c_i}\in \mathcal{S}} \Pr(e_{c_i} | \mathcal{H}).
	\end{equation}
	
	\subsubsection{Variants}
	\label{sec:variants}
	
	We try three other similarity score functions and three other attention functions on the proposed model.
	The score functions are:
	\begin{itemize}
		\item \textbf{C-Score} is the cosine similarity:
		\begin{equation}
			\text{C-Score}(\mathbf{h}_{i}, \mathbf{h}_c) = \frac{\mathbf{h}_{i}^T \mathbf{h}_c}{||\mathbf{h}_{i}|| ||\mathbf{h}_c||} .
			\label{sim:cosine}
		\end{equation}
		
		\item \textbf{M-Score} is the negative Manhattan distance:
		\begin{equation}
			\text{M-Score}(\mathbf{h}_{i}, \mathbf{h}_c) = -||\mathbf{h}_{i} - \mathbf{h}_c||_1 .
			\label{sim:manhattan}
		\end{equation}
		
		\item \textbf{L-Score} is the linear transformation score:
		\begin{equation}
			\!\!\!\text{L-Score}(\mathbf{h}_{i}, \mathbf{h}_c) \!=\! \mathbf{w}\!_{se}^T \mathbf{h}_{i} + \mathbf{w}\!_{sc}^T \mathbf{h}_c + b_{s},\!\!
			\label{sim:linear}
		\end{equation}
		where $\mathbf{w}_{se}, \mathbf{w}_{sc} \in \mathbb{R}^{d_e}$ are the weight vectors and $b_{s} \in \mathbb{R}$ is the bias.
	\end{itemize}
	The attention functions are:
	\begin{itemize}
		\item \textbf{Average attention} simply averages the event-level scores;
		
		\item \textbf{Dot product attention} is calculated by:
		\begin{equation}
			u_{i} = \mathbf{h}_{i}^T \mathbf{h}_c .
			\label{att:dot}
		\end{equation}
		
		\item \textbf{Additive attention} is calculated by:
		\begin{equation}
			u_{i} = \mathbf{w}_{ae}^{T}\mathbf{h}_{i}+\mathbf{w}_{ac}^{T}\mathbf{h}_{c} + b_a ,
			\label{att:additive}
		\end{equation}
		where $\mathbf{w}_{ae}, \mathbf{w}_{ac}\in \mathbb{R}^{d_e}$ are the weight vectors and $b_a\in \mathbb{R}$ is the bias.
	\end{itemize}
	
	\subsection{Training Details}
	
	The training objective is to minimize the cross-entropy loss:
	\begin{equation}
		L(\Theta) = -\frac{1}{N}\sum_{i=1}^{N} \log \Pr(e^*_i | \mathcal{H}_i) + \lambda \text{L}_2(\Theta),
	\end{equation}
	where $e^*_i$ and $\mathcal{H}_i$ denote the correct answer and the historical events for the $i$-th training sample, respectively; $N$ is the number of training samples; $\Theta$ denotes all the model parameters; $\text{L}_2(\cdot)$ denotes the $\text{L}_2$ regularization; $\lambda$ denotes the regularization factor.
	We then train the model using Adam~\cite{DBLP:journals/corr/KingmaB14} algorithm with 100-size mini-batch.
	
	\section{Experiments}
	
	In this section, we compare MCPredictor with a number of baselines to validate its effectiveness.
	In addition, we investigate the variants of MCPredictor and the importance of different constituents.
	Finally we conduct case studies on the model.
	
	\subsection{Dataset}
	
	Following \cite{AAAI1611995}, we extract events from the widely-used New York Time portion of the Gigaword corpus~\cite{graff2003english}.
	The C\&C tool~\cite{curran-etal-2007-linguistically} is used for POS tagging and dependency parsing, and OpenNLP~\footnote{http://opennlp.apache.org} for coreference resolution.
	For the participants that are coreference entities, we select the events they participate in to construct the event chain. For the non-coreference entities, we select the events by matching their headwords. The short chains will be padded with null events (verb and arguments are all null) to the max sequence length.
	Since the development and test sets are unavailable, we generate them using the codes from \citet{AAAI1611995} and their development and test document split.
	Thus, the results of the baseline models may be different from what they are reported in previous studies.
	Detailed information about the dataset is shown in Table~\ref{tab:dataset},
	where \#X denotes the size of X.
	Each sample in the training set, development set, and test set contains five candidate events, where only one choice is correct.
	
	\subsection{Experiment Settings}
	
	\begin{table}
		\centering
		\begin{tabular}{|lc|}
			\hline
			\# train documents	&	830,645	\\
			\# development documents	&	103,583	\\
			\# test documents	&	103,805	\\
			\# train samples	&	1,440,295	\\
			\# development samples	&	10,000	\\
			\# test samples	&	10,000	\\
			\hline
		\end{tabular}
		\caption{Detailed information about the dataset.}
		\label{tab:dataset}
	\end{table}
	
	Following the convention, 
	event sequence length $n$ is set to 8;
	word embedding size $d_w$ is set to 300;
	event embedding size $d_e$ is set to 128;
	the number of Transformer layers is selected from \{1, \underline{2}\};
	the dimension of the feed-forward network in Transformer is selected from \{512, \underline{1024}\};
	the dropout rate is set to 0.1;
	the learning rate is set to 1e-4 except that the learning rate of BERT-tiny  is set to 1e-5;
	the regularization factor $\lambda$ is set to 1e-6.
	All the hyper-parameters are tuned on the development set, and the best settings are underlined.
	All the experiments are run under Tesla V100.
	
	\subsection{Baselines}
	
	\begin{itemize}
		\item \textbf{PMI} \cite{chambers-jurafsky-2008-unsupervised} uses PMI to measure the pairwise similarity of events.
		
		\item \textbf{Event-Comp} \cite{AAAI1611995} uses multi-layer perceptron to encode events and calculates their pairwise similarity.
		
		\item \textbf{PairLSTM} \cite{wang-etal-2017-integrating} uses LSTM to integrate temporal order information into event embeddings.
		
		\item \textbf{SGNN} \cite{ijcai2018-584} merges all the narrative event chains into a narrative event evolution graph.
		The scaled graph neural network is used to obtain graph information in event embeddings.
		
		\item \textbf{SAM-Net} \cite{DBLP:conf/aaai/LvQHHH19} uses an LSTM network and self-attention mechanism to capture semantic features.
		
		\item \textbf{NG}  \cite{lee-etal-2020-weakly} builds a narrative graph to enrich the event semantic representation via introducing discourse relations.
		
		\item \textbf{Lv2020} \cite{lv-etal-2020-integrating} introduces external commonsense knowledge and uses a BERT-based model to predict the subsequent event.
		
		\item \textbf{SCPredictor}, an ablation of MCPredictor, removes the scores from the other narrative event chains at the script level.
		
		\item \textbf{SCPredictor-s and MCPredictor-s}, ablations of SCPredictor and MCPredictor, respectively, remove the sentence information at the event level.
		
	\end{itemize}
	
	For PMI, we use the version implemented by \cite{AAAI1611995}.
	For PairLSTM, we adopt the version implemented by \cite{ijcai2018-584}.
	For others, we use the versions provided in their original papers.
	For comparison, accuracy(\%) of predicting the correct subsequent events is used as the evaluation metric.
	
	\subsection{Results and Analyses}
	
	\begin{table}
		\centering
		\begin{tabular}{lc}
			\hline
			Method        &  Accuracy(\%)  \\ \hline
			PMI*          &     30.52      \\
			Event-Comp*   &     49.57      \\
			PairLSTM*     &     50.83      \\
			SGNN*         &     52.45      \\
			SAM-Net*      &     54.48      \\
			Lv2020*       &     58.66      \\
			NG*           &     63.59      \\ \hline
			PMI           &     30.28      \\
			Event-Comp    &     50.19      \\
			PairLSTM      &     50.32      \\
			SGNN          &     52.30      \\
			SAM-Net       &     55.60      \\ \hline
			SCPredictor-s &     58.28      \\
			MCPredictor-s &     59.24      \\
			SCPredictor   &     66.24      \\
			MCPredictor   & \textbf{67.05} \\ \hline
		\end{tabular}
		\caption{Model accuracy on the test set. ``*'' denotes the performance from previous studies.}
		\label{tab:result}
	\end{table}
	
	Final experimental results on the script event prediction task are shown in Table~\ref{tab:result}.
	From the results, we have the following observations:
	
	\begin{itemize}
		\item SCPredictor and MCPredictor outperform SCPredictor-s and MCPredictor-s respectively by more than 7.81\%.
		This significant improvement indicates that introducing sentence information at the event level successfully enhances the event representations;
		
		\item MCPredictor-s and MCPredictor outperform SCPredictor-s and SCPredictor respectively, 
		which demonstrates the effectiveness of introducing multiple narrative event chains at the script level.
		In addition, the improvement of multiple chains decreases slightly when the sentence information is introduced (from 0.96\% to 0.81\%).
		This is probably because extra events brought by other narrative event chains may partially covered by the sentences corresponding to existing events.
		In the development set, only about 13\% of the samples contain the extra events that are not covered by the sentences corresponding to existing events when introducing other narrative event chains.
		Still, the multiple narrative event chains contribute to the model;
		
		\item SCPredictor-s outperforms existing models (i.e., PMI, Event-Comp, PairLSTM, SGNN, and SAM-Net) by more than 2.68\% under the same input (only the verb and the headwords of the participants are used).
		This improvement indicates that the Transformer networks can model the script better than existing network structures.
	\end{itemize}
	
	\subsection{Human Evaluation}
	
	In addition to automatic evaluation,
	we also conduct human evaluation to further study the performance of the proposed MCPredictor model.
	Specifically, we randomly select 100 samples in the development set.
	Without loss of generality, we select SAM-Net model to compare with MCPredictor manually under these samples.
	MCPredictor correctly answers 72 out of 100,
	while SAM-Net only correctly answers 60.
	These results demonstrate that the MCPredictor generally derives more plausible answers.
	
	\subsection{Comparative Studies on Variants}
	
	\begin{table}
		\centering
		\begin{tabular}{lcc}
			\hline
			Method  & Accuracy(\%) & $\Delta$ \\ \hline
			E-Score &    67.35     &    /     \\
			M-Score &    65.36     &  -1.99   \\
			L-Score &    65.05     &  -2.30   \\
			C-Score &    60.68     &  -6.63   \\ \hline
		\end{tabular}
		\caption{Performance comparison of MCPredictor over different scoring functions.}
		\label{tab:scoring}
	\end{table}
	
	\begin{table}
		\centering
		\begin{tabular}{lcc}
			\hline
			Method     & Accuracy(\%) & $\Delta$ \\ \hline
			scaled-dot &    67.35     &    /     \\
			dot        &    66.21     &  -1.14   \\
			additive   &    65.84     &  -1.51   \\
			avg        &    64.74     &  -2.61   \\ \hline
		\end{tabular}
		\caption{Performance comparison of MCPredictor over different attention functions.}
		\label{tab:attention}
	\end{table}
	
	To further study the effects of different scoring functions and attention functions, 
	we conduct comparative studies on the development set.
	
	The comparative results on different scoring functions are listed in Table~\ref{tab:scoring},
	where $\Delta$ means the performance difference between the best scoring function and each alternative scoring function.
	As presented in Table~\ref{tab:scoring},
	E-Score achieves the best performance among the four scoring functions,
	which is consistent with the result in \cite{ijcai2018-584}.
	The reason for C-Score underperforming the other three scoring functions may be that it only measures the angle between the two vectors, however, ignores their lengths.
	
	Similarly, the comparative results on different attention functions are listed in Table~\ref{tab:attention}.
	In this table, scaled-dot is the scaled-dot product attention in Equation~\ref{att:scaled-dot},
	dot indicates the dot product attention in Equation~\ref{att:dot},
	additive is the additive attention in Equation~\ref{att:additive},
	and avg denotes the average attention.
	Similar to Table~\ref{tab:scoring}, $\Delta$ means the performance difference.
	As shown in Table~\ref{tab:attention}, the scaled-dot product attention outperforms the others.
	The average attention underperforms the other attention functions,
	which shows the effectiveness of the attention mechanism.
	The reason for dot product attention under-performing the scaled-dot attention function may be that dot product attention tends to give a single event too much weight,
	which may suppress the contributions of other related events.

	\subsection{Detailed Analyses on Constituents}
	
	\begin{table}
		\centering
		\begin{tabular}{lcc}
			\hline
			Constituents & Accuracy(\%) & $\Delta$ \\ \hline
			All          &    67.35     &    /     \\
			-V           &    60.54     &  -6.81   \\
			-N           &    64.81     &  -2.54   \\
			-J           &    66.89     &  -0.46   \\
			-R           &    67.03     &  -0.32   \\
			-V(self)     &    67.29     &  -0.06   \\
			-V(others)   &    60.90     &  -6.45   \\
			-V\&N         &    55.36     &  -11.99  \\
			-V\&R         &    60.23     &  -7.12   \\
			-N\&J         &    64.10     &  -3.25   \\
			-R\&J         &    66.55     &  -0.80   \\ \hline
		\end{tabular}
		\caption{Performance comparison of MCPredictor over different constituents.}
		\label{tab:constituents}
	\end{table}
	
	\begin{table*}
		\centering
		\begin{tabular}{cc}
			\hline
			Protagonist A: First Virtual	&	Original Sentence	\\
			\hline
			\textbf{set+up(executive, company)}	&	First Virtual, which Ungermann \textit{set up} in ...	\\
			\textbf{demonstrate(executive, product)}	&	First Virtual ... publicly \textit{demonstrated} its products ...	\\
			\textbf{aim(company, null, segment)}	&	First Virtual is \textit{aiming} at an emerging segment ...	\\
			\textbf{be(company, able)}	&	\multirow{3}{3in}{First Virtual \textit{is} able to \textit{offer} its lower prices by \textit{limiting} its ATM network speeds to 25 million bits a second ...}	\\
			\textbf{offer(company, price)}	&		\\
			\textbf{limit(company, speed, bit)}	&		\\
			\textbf{sneak(company, null, organization)}	&	it could \textit{sneak} into organizations ...	\\
			\textbf{develop(company, software)}	&	First Virtual has also \textit{developed} software ...	\\
			\hline
			protagonist B: Ralph Ungermann	&	Original Sentence	\\
			\hline
			consider(father, executive)	&	Ralph Ungermann ... is \textit{considered} a founding father ...	\\
			set+out(executive, problem)	&	Ralph Ungermann ... has \textit{set out} to solve his newest company ...	\\
			set+up(executive, company)	&	First Virtual, which Ungermann \textit{set up} in ...	\\
			predict(executive)	&	Ungermann \textit{predicts} that ...	\\
			contend(executive)	&	Ungermann \textit{contends} that ...	\\
			\hline
			\multicolumn{2}{c}{correct choice: sell(executive, company); wrong choice: launch(company, attack)}	\\
			\hline
		\end{tabular}
		\caption{Case study of MCPredictor. The historical events used by SCPredictor are in boldface.}
		\label{tab:case_multichain}
	\end{table*}
	
	What is the contribution of each constituent from the sentences?
	To go deep into it, we use a masking mechanism to hide some constituents.
	The results on the development set are listed in Table~\ref{tab:constituents},
	where ``All'' denotes that all constituents are used;
	``V'' denotes the verbs;
	``N'' denotes the nouns;
	``J'' denotes the adjectives;
	``R'' denotes the adverbs;
	and ``-X'' denotes that X is masked.
	In addition, we study the influence of their pairwise combinations X and Y (denoted as ``X\&Y'').
	Considering the modification relationship between the constituents,
	we only conduct the experiments on the combinations of ``V\&R'', ``V\&N'',
	``N\&J'', and ``R\&J'', respectively.
	
	Masking the verbs brings 6.81\% decrease to the performance.
	To further study the influences of different verbs,
	we mask the focused verb (denoted as ``V(self)'') and the verbs excepting the focused one (denoted as ``V(others)'') separately.
	The results show that masking the focused verb almost do no harm to the performance,
	while masking others brings 6.45\% decrease to the performance.
	This is because the event encoding component still obtains the information about the focused event despite the text encoding component masking it.
	In addition, these results show that the implicit relevance of the other verbs and the focused verb is important for script event prediction.
	
	Masking the nouns brings 2.54\% decrease to the performance.
	The nouns are usually the participants
	which influence the word sense of the verbs.
	Thus, the nouns are also important for prediction.
	
	Masking the adjectives brings 0.46\% decrease to the performance when comparing ``All'' and ``-J'',
	while it brings 0.71\% decrease when comparing ``-N'' and ``-N\&J''.
	This phenomenon shows that using the combination of the nouns and the adjectives brings richer semantic information than using them separately.
	The same phenomenon also appears when comparing ``All'', ``-V'', and ``-V\&R'' .
	
	\subsection{Case Studies}
	
	To have a better understanding of the effects of multiple narrative event chains, 
	we study the cases in the development set where the MCPredictor model selects the correct choice while the SCPredictor model selects the wrong one.
	As presented in Table~\ref{tab:case_multichain},
	protagonist A is ``company'' in the events, and protagonist B is ``executive'' in the events.
	When only the narrative event chain of protagonist A is provided,
	the correct candidate event is not very convincing,
	since protagonist B is rarely mentioned in this chain.
	In addition, the events in this chain are common, and many events can be the subsequent event.
	Thus, SCPredictor is likely to select the wrong candidate event as the prediction result.
	On the contrary, MCPredictor integrates evidence from both narrative event chains.
	It is thus able to predict the correct candidate event.
	Moreover, the events contain less information comparing to the original sentences,
	which shows the necessity to introduce the sentence information for them.
	
	\section{Conclusion and Future Work}
	
	In this paper, we proposed the MCPredictor model to handle the script event prediction task,
	which integrates deep event-level and script-level information.
	At the event level, MCPredictor additionally utilizes the original sentence to obtain more informative event representation.
	At the script level,
	it considers multiple event sequences corresponding to different participants of the subsequent event.
	Experimental results demonstrate its merits and superiority.
	
	However, currently, we still need to know the participants of the candidate events to extract multiple chains.
	When participants are unknown, we have to enumerate all possible combinations of entities in the script, which is time-consuming.
	Moreover, extracting informative constituents of events besides the verb and their participants is still a challenge.
	In the future, we will study these problems.
	
	\section*{Acknowledgment}

	The work is supported by 
	the National Natural Science Foundation of China under grants U1911401, 62002341, 61772501, U1836206 and 61722211, 
	the GFKJ Innovation Program, 
	Beijing Academy of Artificial Intelligence under grant BAAI2019ZD0306, 
	the Lenovo-CAS Joint Lab Youth Scientist Project,
	and the Foundation and Frontier Research Key Program of Chongqing Science and Technology Commission (No. cstc2017jcyjBX0059).
	Thanks to Yixing Fan, Huawei Shen, Jinhua Gao, Yunqi Qiu, Xiao Ding, and the reviewers for the constructive discussions and suggestions.
	
	\bibliography{anthology,custom}
	\bibliographystyle{acl_natbib}
	
\end{document}